# Curve Trajectory Model for Human Preferred Path Planning of Automated Vehicles

Gergo Ferenc Igneczi*[1], Erno Horvath[2], Roland Toth[3] and Krisztian Nyilas[4]

[1] Affiliation: Vehicle Research Center, Szechenyi Istvan University, Egyetem ter 1, Gyor, Hungary, 9026; personal contact information: email: gergo.igneczi@ga.sze.hu, +36-70-403-57-39, orcid: 0000-0002-1258-837X
[2] Affiliation: Vehicle Research Center, Szechenyi Istvan University, Egyetem ter 1, Gyor, Hungary, 9026; personal contact information: herno@ga.sze.hu, orcid: 0000-0001-5083-2073
[3] Affiliation: Institute for Computer Science and Control, Kende str. 13-17., Budapest. Hungary, 1111; personal contact information: email: toth.roland@sztaki.hu, orcid: 0000-0001-7570-6129
[4] Affiliation: Robert Bosch Kft, Gyomroi str. 104 - 120, Budapest, Hungary, 1103; personal contact information: email: Krisztian.Nyilas@hu.bosch.com
*: corresponding author

## Acknowledgement

The research was supported by the European Union within the framework of the National Laboratory for Autonomous Systems. (RRF-2.3.1-21-2022-00002).

# Curve Trajectory Model for Human Preferred Path Planning of Automated Vehicles

**Abstract**
Automated driving systems are often used for lane keeping tasks. By these systems, a local path is planned ahead of the vehicle. However, these paths are often found unnatural by human drivers. We propose a linear driver model, which can calculate node points that reflect the preferences of human drivers and based on these node points a human driver preferred motion path can be designed for autonomous driving. The model input is the road curvature. We apply this model to a self-developed Euler-curve-based curve fitting algorithm. Through a case study, we show that the model based planned path can reproduce the average behavior of human curve path selection. We analyze the performance of the proposed model through statistical analysis that shows the validity of the captured relations.

## 1 Article Highlights

- We have identified the behavior of human drivers for selecting lateral offsets w.r.t. the middle of the lane while driving in curves.
- We have built a model which can reproduce human-like path geometries.

## 2 Introduction

Development of Advanced Driver Assistance Systems (ADAS) has accelerated in the last decade. Next to premium brands, more and more vehicle manufacturers decide to develop and sell automated driving functions for safety and comfort. The European Union has made automated emergency braking and emergency lane keep assist functions mandatory from 2022. The openly available NCAP tests include the rating of active safety functions for almost a decade. People travel more year-by-year, either due to working purposes or for free time activities. This clearly motivates the further development of automated driving functionalities.

We usually distinguish between longitudinal and lateral control functions. Longitudinal functions are realized by Adaptive Cruise Control (ACC). Many ACC systems are capable of driving in high traffic density in the speed range of 0 to 200 *kph*, including stop-and-go functionality. Lateral functions are usually called Lane Keep Assist (LKA). This function can be activated automatically for emergency situations (only when vehicle crosses the lane edge), or can be continuously active, guiding the vehicle within the lane. The first one is usually referred as Emergency Lane Keep Assist, while the latter one as Active Lane Driving. In this paper, we focus on the lateral control system, therefore in the sequel we concentrate on the lateral control components.

ADAS architecture has not changed significantly for decades. The main components of ADAS are detection, perception, behavior planning, motion planning and actuator control. Our work focuses on the motion planning component. We assume that the preliminary components from detection to behavior planning are available and work reliably. As we have shown in our previous work [25], there are multiple concepts on defining the motion planning problem. The most used approach in automotive industrial solutions is the corridor-based approach, where the corridor forms a free space ahead of the vehicle. The corridor is defined by the lane edges as borders. The task is to plan a local path within the corridor based on safety and comfort aspects. The path must be kinematically feasible considering the vehicle control constraints.

## 3 Problem Formulation

One of the key tasks within the automated driving systems is to define a local trajectory, which describes the target path in front of the vehicle. Such a trajectory planning usually stands for designing a geometric target path and a respective kinematic profile (i.e., speed profile). Often the geometric target path is simplified to follow the midline of the lane. This is unnaturalistic, as human drivers, besides of driving on long straight roads, often do not follow such a strategy.

In the literature there are already approaches to combine various objectives (e.g., comfort, safety, efficiency) and human-likeness in trajectory planning. Several works use optimization techniques to provide trajectories that are optimal based on the above objectives [1-3, 8, 16-18], or do the same to generate optimal control trajectories [6-7]. These approaches rely on assumption that human drivers also prefer to optimize such factors. However, optimizing on abstract kinematic quantities can be difficult in real time, and do not implicitly provide human-likeness. There are various solutions using machine learning algorithms [4-5, 9, 11]. However, these are quite complex algorithms, which consume much computational power.

The research question we intend to investigate is how we can produce a robust driver model which can plan human-like path geometries. We neglect kinematic effects in the sequel and will only include them in the modelling problem in our future work.

The paper structure is the following: first, we give an overview on existing solutions for such a path planning problem. Then, we introduce our observations based on real-world driving data, and we propose a driver model structure. Then, we discuss the implementation and calibration of the driver model. Finally, we make validation tests through statistical analysis of the data and a case study.

## 4 Literature Overview

In the field of driver models for path planning, there are only limited number of works in the literature. There are many driver models for motion controllers [10, 15, 23-24]. These driver models focus on how drivers act on the actuators to realize motion targets. However, they do not cover how the target path is generated.

Driver trajectory models are proposed which aim to plan trajectories that are preferred by human drivers. However, these models often neglect the human-likeness of the path and optimize abstract kinematic quantities of the trajectory. There are machine learning techniques that are used to generate human-like paths. Inverse Reinforcement Learning (IRL) method is used to train the model on human-like path set [11]. The proposed reward function of the IRL has four terms for both longitudinal and lateral planning: velocity keeping, lane keeping, lane boundary keeping and collision avoidance. This solution gives a good result in complex traffic scenarios, but still misses the essential path selection preferences of the drivers. The lane keeping term calculates an error between the vehicle and the lane center. However, the lane center is usually not the preferred path by human drivers.

There is a trajectory planner using multiple Euler-curves optimized on lateral jerk and acceleration to design a curve trajectory [17]. By minimizing the lateral acceleration and jerk, this trajectory will provide high level comfort for passengers. However, this solution lacks the connection to actual human chosen paths and rely on purely the fact that minimization of acceleration and jerk is sufficient to meet passengers' requirements of smooth driving. Similar optimization criteria are used for [8]. A path model is proposed which has a smoothing spline to reduce lateral jerk and acceleration in curve transitions [18]. The article states the smoothing is a sufficient criterium of human-like path selection.

A method is introduced to generate a global trajectory which is optimized for racing [12]. The aim is to reduce the lap time by increasing the curve speed of the vehicle. The article states that path curvature minimization is the ultimate way to reduce the centripetal force on the vehicle, and thus maximizing curve speed. This is an important statement as it indicates curvature is one of the key factors influencing the trajectory shape.

There are only a few works which provide planning solutions for human-like geometry of the path. In [16], it has been shown how human like trajectories can be generated for lane change maneuvers. The planner provides a combined solution, where optimization on different cost terms, e.g., comfort, stability and dynamics happens. However, an additional term is added to the cost function which is associated with the distance error between the drivers' path and the planned path. The cost weights are learnt by optimization on a reference data set. This produces human-like path shapes through optimization and learning.

In [13], a driver path model is introduced, which is trained on real-world driving data. The article has shown that there is a linear relation between the lane offset selection and the lane offsets from the previous 2 seconds. There is a proposed linear model with the structure in Equation (1).

$$D_i = a_{i-1}D_{i-1} + a_{i-2}D_{i-2} + a_{i-3} + e_i, \qquad (1)$$

where $i$ is the calculation cycle, $a_{i-1}$, $a_{i-2}$ and $a_{i-3}$ are the regression coefficients, $D_i$, $D_{i-1}$, $D_{i-2}$ are the lane offsets and $e_i$ is a random error. The coefficients were calculated based on the data of 30 drivers in one given curve. This article has shown that drivers tend to select lane offsets based on two consecutive points in a recursive manner. Based on these results, we can take the assumption that the lane offset change in three consecutive points ($D_{i-2}$, $D_{i-1}$ and $D_i$) can reflect the orientation change, which is directly connected to the curvature. Therefore, curvature may play an important role in the definition of the lane offset chosen by drivers.

We can see from the literature overview, that there are usually two approaches for human-like path planning:
- trajectory planning based on optimization of various kinematic quantities, which eventually produce a path which is close to driver selected path in its geometry, or
- direct path planning approaches, where perceived information is used to produce a path which preserves the geometry preferences of human drivers. These solutions usually give a direct relation between human-likeness and the resulting path. Hence, they provide a more accurate replicate of human path planning. However, there are only a few works available for this selection approach.

To our best knowledge, there are no existing solution to model the base behavior of human path planning. With our approach we provide a model to describe the geometric planning strategy of human drivers.

## 5 Driver Model

In this Section we introduce our assumptions based on which the driver model structure is formulated. Then, we provide information of the data that is used for analysis. Our observations are detailed, then the model structure is set up based on these observations.

### 5.1 Model System

In our work we focus on providing a path planning model which can be used to reproduce human-like path geometries. Our model system relies on the following assumptions:
- Assumption 1: Drivers follow a planning – control approach.
- Assumption 2: Drivers perceive environmental information, although they do not use all the information, but only information from nominated road points (node point model).

- Assumption 3: Between nominated road points they follow a preconditioned behavior (i.e., one given curve type – curve fitting model).
- Assumption 4: They plan their behavior based on the perceived information (offset model).
- Assumption 5: They do replanning, if needed (retrigger model).

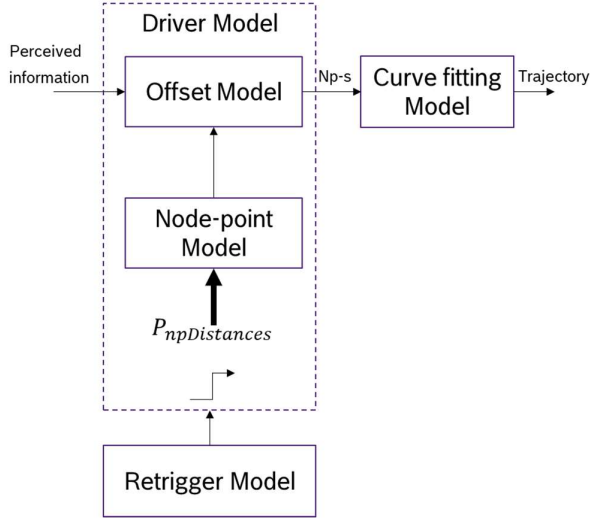

**Fig.1** Proposed model system for the curve path model

The proposed model system can be seen in Fig.1. We divide the problem into four main parts. Based on our second assumption, drivers perceive information within a preview distance, but we condense the total information into node points. The number of points may vary. The model that produces the node points is called the node point model. Based on our third assumption, drivers plan their behavior based on the condensed information and then follow the planned path according to instinctive preferences (comfort, speed, etc.). Therefore, we introduce a curve fitting model, which fits a human-preferred shaped curve onto the node points. According to our fourth assumption, drivers plan their behavior based on what they perceive from the preview distance. This can be simplified to planning the offsets compared to the mid-lane in the node points. We call this an offset model.

Using our fifth assumption, drivers replan their path occasionally. This is what we call a retrigger model.

In the followings, we provide details of the data and our observations to identify the structure of the model elements.

### 5.2 Data Analysis

The aim of our work to identify a driver model based on real-world driving data, which can replicate human-like path geometry. As seen in Fig.1, we divide the driver model system into model elements. The two core elements of the driver model are the node point model and the offset model. To set up the structure of these models, the data of 15 drivers have been collected. Each driver drove approximately 40 $km$ in real-world traffic. The test route lies on a two-lane road, (Main Road 62), between cities of Dunaújváros and Székesfehérvár, in Hungary. The map is given in Fig.2.

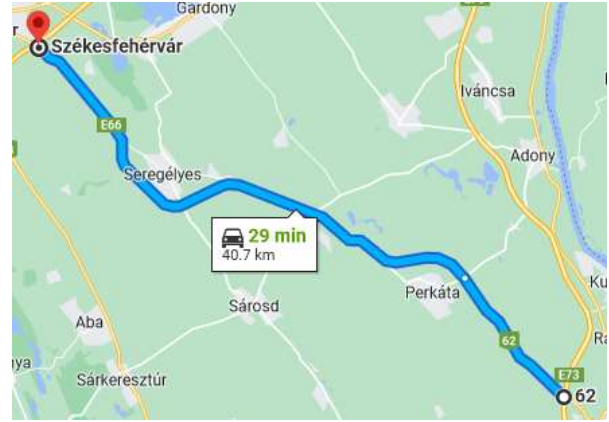

**Fig.2** The test route is Main Road 62 between cities of Dunaújváros and Székesfehérvár in Hungary

The test vehicle has been a Skoda Octavia MK3 with automatic gearbox. The measurements took place during workdays between 10AM and 2PM. The weather was always dry, visibility conditions were appropriate. There was no high traffic density nor traffic jams.

The vehicle has been equipped with a lane detection system, which measures the road geometry. The lane edges define the shape of the road. These edges are represented by a third order polynomial in our data set. The path at time point $t$ can be given in Equation (2).

$$\bar{y}_i = c_0^i + c_1^i \bar{x}_i + 2c_2^i \bar{x}_i + 6c_3^i \bar{x}_i, \qquad (2)$$

where $c_0$ is the lateral distance, $c_1$ is the orientation, $c_2$ is the curvature and $c_3$ is the curvature change of the lane, $\bar{x}_i$ are the longitudinal coordinates and $\bar{y}_i$ are the lateral coordinates of the foreseen road mid-lane in the planning frame, at calculation cycle $i$. The camera system has a look ahead distance of 150 $m$.

We use the following sign conventions:
- positive offset – left side of mid-lane
- negative offset – right side of mid-lane
- positive curvature – left curve
- negative curvature – right curve

### 5.3 Observations

We assume, that drivers have different path selection behavior when driving in straight lines and curves. The key variable in the modelling problem is therefore assumed to be the road curvature. Similar statement has been made in other articles [12-13, 16]. The road curvature is directly measured by our lane detection system.

In Fig.3 we can see the relation between road curvature and selected driver offset. Out of the test drivers several drivers showed the same behavior. The offset seems to be dependent on the road curvature. The relation is close to linear: the higher the curvature is, the higher the selected offset to the mid-lane is. This indicates a curve cutting behavior.

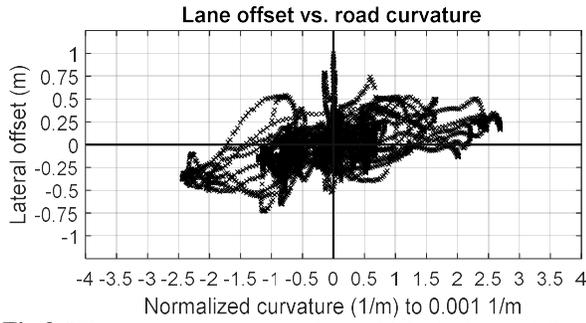

**Fig.3** Offset - curvature plot of a driver. Multiple drivers follow the same characteristic. The entire route data of Driver 1 has been plotted. Curvature is positive in left curves, negative in right curves. Offset to the mid-lane is positive when the vehicle is on its left side, and negative on the right side

However, we can also see, that for one given curvature multiple offset points are taken by the driver. This can be due to the fact that the figure contains the data of the entire test route. On the other hand, we can also assume, that a scalar function between the road curvature and the lane offset cannot appropriately describe the path planning problem. Therefore, we aim to use a multi-dimensional model which can produce such non-scalar function behavior, and as such, the number of node points will be higher than one. We use the road curvature as input, and the lane offset as the output of the model. The modelling problem is formulated in Equation (3).

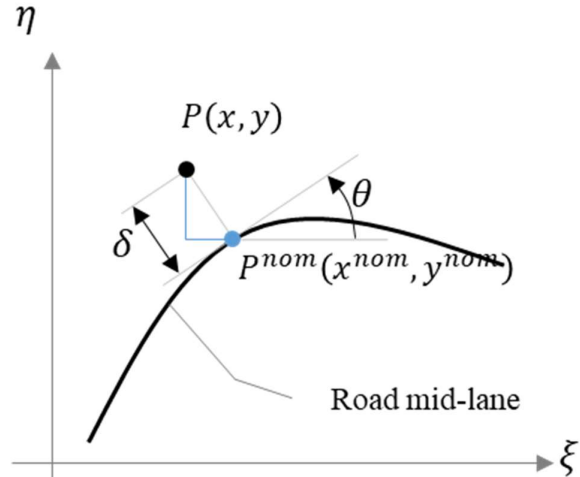

**Fig.4** Visualization of road offset value

$$y(\xi) = y^{nom}(\xi) + \delta(\kappa)\cos(\theta(\xi)),$$
$$x(\xi) = x^{nom}(\xi) - \delta(\kappa)\sin(\theta(\xi)),$$
$$\delta(\kappa) = f(\kappa(\xi)), \qquad (3)$$

where:
- $[\xi, \eta]$ is a freely chosen coordinate frame,
- $[x^{nom}, y^{nom}]$ defines the nominal node point on the middle of the road,
- $[x, y]$ defines the final node point corrected with the offset value,
- $\delta$ is the offset value,
- $\theta$ is the road orientation angle,
- $\kappa$ is the road curvature,
- $f(\kappa(\xi))$ is the model function.

The above quantities are visualized in Fig.4.

### 5.4 Model Structure

Based on the aforementioned observation, we propose the following model structure for the node point model and the offset model:
- nominate three node points within the maximum preview distance (node point model).
- Use road curvature as the key variable in the modelling problem.
- Calculate the offset to the mid-lane in the node points (offset model).

When selecting the number of node points, it is important to minimize the complexity of the algorithm, which makes it attractive to choose low number of node points. On the other hand, we would like to provide enough flexibility for the model, therefore choosing a relatively high number of node points is more advantageous. We have also seen in the previous chapter, that at minimum two node points are necessary. Various driver models in the motion control field use look ahead point based approaches [10, 15, 23-24]. These often use one or two look ahead points. As we have a long preview distance by the camera system (150 m), we extend the number of node points to three. We believe that three node points within the preview distance can provide enough flexibility and yet avoid high calculation complexity.

A trade-off analysis on selecting different number of node points have been done. We have simulated one route section and set up the system to fit a curve on node points lying on the mid-lane of the road. This way we exclude the error coming from the variance of the human offset selection. Two indicators were calculated:
- mean error distance between points of the planned path and the mid-lane as reference,
- mean time of the planning sequence.

One planning step happens at a retrigger, and includes node point distance calculation, offset calculation and curve fitting, while the planning sequence means all planning steps during the road. The number of node points have been varied between 1 and 10. Then, both indicators have been normalized by their maximum value through the node point range. The results can be seen in Fig.5. The results show that the calculation time starts increasing after 4 node points significantly, while the mean error distance improves less and less with increasing number of node points. Based on this observation we have chosen three node points in our simulation.

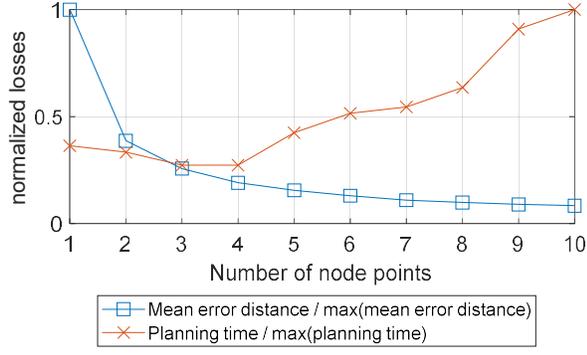

**Fig.5** Loss vs. calculation time of planning in the case of different number of node points

We will call the three node points the near, mid and far range points, accordingly. The node point distances are provided as parameters of the model and can be seen in Equation (4).

$$P_{npModel} = [d_n\ d_m\ d_f]. \tag{4}$$

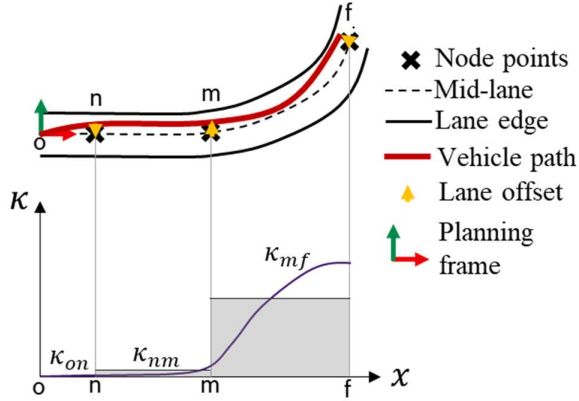

**Fig.6** Illustration of one planning cycle in a curve entry situation. This is a generic example of the node point, curvature vector and offset vector selection

According to our second assumption, drivers perceive environmental information, but they consider only the information about the nominated road points when making decisions. Considering three node points and the road curvature as the input of our model, we propose to take the average value of the road curvature between the node points. The three node points and the origin of the planning frame yield three subsections. Thus, we have three average curvature values. For planning cycle $i$, the curvature input vector is the following:

$$\bar{\kappa}_i = \begin{bmatrix} \kappa_{on_i} \\ \kappa_{nm_i} \\ \kappa_{mf_i} \end{bmatrix}, \tag{5}$$

where $on$, $nm$, $mf$ stand for origin-near, near-mid and mid-far subsection, respectively.
Based on our fourth assumption, drivers plan their behavior using the perceived information. This is what we call an offset model. Based on our Equation (3), we calculate the offsets in the node points compared to the mid-lane. In planning cycle $i$, this yields the following model output vector:

$$\bar{\delta}_i = \begin{bmatrix} \delta_{n_i} \\ \delta_{m_i} \\ \delta_{f_i} \end{bmatrix}, \tag{6}$$

where $n$, $m$, $f$ stand for near, mid and far range points.
We can see the illustration of a generic example in Fig.6. This is a curve entry scenario, where the planning frame is positioned in the straight section before the curve. In the preview distance, the curvature starts to increase. The node point offsets are always calculated as a perpendicular distance to the road mid-lane in the given node points.

In Section 5.2, we have observed that there is a linear relation between the road curvature and the selected lane offset. Therefore, we propose a linear offset model, which calculates node point offset values based on the subsegment average curvature data. We call the model Linear Driver Model (LDM). Equation (3) model function is given for the $i^{th}$ calculation cycle in Equation (7).

$$\bar{\delta}_i = f(\bar{\kappa}_i) = P\bar{\kappa}_i, \tag{7}$$

where:
- $P \in \mathbb{R}^{3x3}$ is the parameter matrix that contains the weights of the linear combination of the curvature values,
- $\bar{\kappa}_i = \begin{bmatrix} \kappa_{on}\ \kappa_{nm}\ \kappa_{mf} \end{bmatrix}_i^T$ is the input vector in the $i^{th}$ calculation cycle,
- $\kappa_{on}$, $\kappa_{nm}$ and $\kappa_{mf}$ are average curvature values in the subsections $on$, $nm$ and $mf$ respectively.

The combination of the node point model and the offset model formulates the complete driver model, as indicated in Fig.1. The model is parametrized in terms of $P$ and $P_{npModel}$. Our assumption is that by selecting proper parameter values for both node point distances and the offset model, human-like paths can be generated.

### 5.5 Implementation

In this section we propose a method to calibrate the driver model introduced in Section 5.3. So far, we have discussed how the model can be structured to plan human-like path in one given planning cycle. However, for real-time application we must ensure, that the planning occurs continuously while driving. Before providing the calibration for the driver model, we introduce the implementation of the retrigger and the curve fitting model.

Based on our fifth assumption, drivers do replan occasionally. This is called a retrigger model. For each replanning cycle, the actual perceived information is used to design the node point offsets. We propose to use cyclic replanning. This means, that the node point calculation and curve fitting occur in even cycles. The retrigger cycle can be varied.

According to our third assumption, drivers plan their behavior based on instinctive preferences between node points. Therefore, we fit a curve onto the node points. In our previous work we have proven, that Euler-curves can describe the geometry of human paths with high accuracy [25]. This is called the curve fitting model.

We fit three Euler-curves onto the node points. One curve is fitted onto each point pairs: $on$, $nm$ and $mf$. The Euler-curve can be fitted by four boundary conditions:
- starting point position and orientation and
- end point position and orientation.

The position conditions are given by the node point positions, while the orientation condition is equal to the respective road orientation. The road orientation is provided by the lane detection system.

The implemented model and planner architecture can be seen in Fig.7. We have implemented the algorithm in MATLAB. Using pre-recorded data of the test drivers, simulation is done. The simulation has two different options:
- Model estimation simulations: the offset model is replaced by measurement data, therefore taking over the offset values selected by the driver,
- Model validation simulations: the offset model is implemented according to Equation (7).

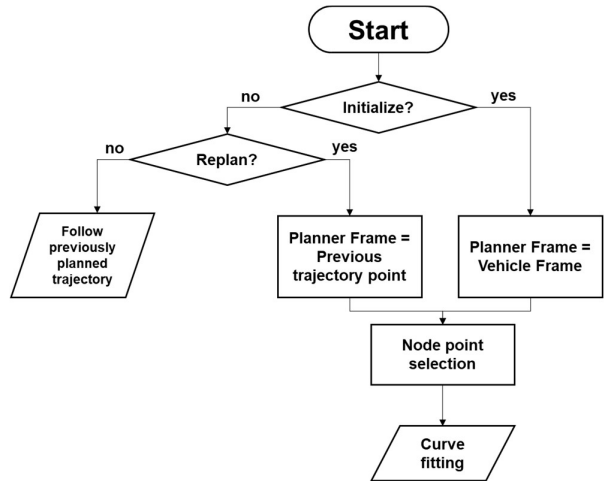

**Fig.7** Implemented architecture of the model and the planning algorithm

The driver model and the retrigger model have the following parameters:
- $P_{npModel} = [d_n\ d_m\ d_f]$ – node point distance from the planner frame,
- $P \in \mathbb{R}^{3 \times 3}$ – the parameter matrix of the offset model,
- $p_{retrigger}$ – the retrigger cycle of the planning.

Next, we propose approaches to give the values of these parameters.

The node point distances can be selected based on various approaches. A relatively simple approach is to select node points within the preview distance which are equidistant to each other. However, such solution would not have any connection to the actual human curves. It has also been shown that a more accurate fit on drivers' paths can be achieved by selecting non equidistant node points. In our previous work, we have shown, that the best fit on human path can be achieved if the node point distances are not equidistant but follow an increasing gap between each other [25]. As a result of an optimization executed on the data of three professional drivers, we have calculated the following node point distances:

$$P_{npModel} = [d_n\ d_m\ d_f] = [10.0\ 39.0\ 137.0]\ m \qquad (8)$$

During the optimization process, we make the following steps:
- we place three equidistant node points within a maximum preview distance of 250 $m$,
- a curve is fitted on the node points, this is the local path,
- the curve points are given in the exact same points where the human paths are recorded,
- the mean distance between the fitted and the recorded path points are calculated, this is the cost of the optimization,
- the node point distances are varied to minimize the cost,
- the exact same procedure is done for the entire test route, with a step size of 400 data samples, which is 8 $s$,
- for each optimization cycle, an optimal distance vector is calculated,
- then, all the distance vectors are averaged to get the overall distance vector, given in Equation (8).

The details of the optimization, the cost calculation, and the results can be found in [25].

The parameter matrix of the offset model can be calculated by linear regression on the human drivers' data. This provides path selection that is human-like. We propose to accomplish linear regression for each driver separately, providing the best fit on personal path selection preferences. For this purpose, model in Equation (7) can be reformulated as

$$D = PU + \varepsilon, \qquad (9)$$

where:
- $\varepsilon$ is the regression error,
- $D = [\overline{\delta}_1\ \overline{\delta}_2\ \cdots\ \overline{\delta}_N]$, $D \in \mathbb{R}^{3 \times N}$ is the node point offset matrix,
- $U = [\overline{\kappa}_1\ \overline{\kappa}_2\ \cdots\ \overline{\kappa}_N]$, $U \in \mathbb{R}^{3 \times N}$ is the input matrix,
- $N$ is the number of calculation cycles.

We search the value of $P$, so that $\varepsilon$ would be minimal. This is equivalent to minimizing the 2 norm of $\varepsilon = D - PU$. By the Moore-Penrose pseudoinverse of $U$, the solution is

$$P = DU^T(UU^T)^{-1}. \qquad (10)$$

The even cycle of replanning is chosen to provide replanning frequently. Between two replanning, the previously planned path is followed. We choose the replanning cycle based on two aspects:

- follow the previously planned path as long as possible to exploit the planning characteristics,
- do replanning as soon as possible to provide flexible reaction on upcoming perception information.

As the nominal vehicle speed in the test route is $25\frac{m}{s}$, the sampling time of the dataset is $0.05\ s$, and considering that the mid-range node point lies $39.0\ m$ from the planning frame, the vehicle reaches the mid-range node point in 30 *cycles*. Therefore, we choose the replanning cycle to 30 *cycles*, which stands for $1.5\ s$.

In the following section, we make simulation runs to validate the model behavior, using the parameters obtained in this section.

## 6 Validation

### 6.1 Validation Concept

The driver model is validated with two different approaches. First, we show the operation of the model through a case study, comparing the human path of one driver and the model output path in terms of lateral offsets and curvature. Secondly, we introduce a comprehensive statistical analysis on the model output, comparing the model to human drivers.

### 6.2 Test Participants

15 drivers have been asked to participate in the driving. Their driving data has been recorded. The drivers are selected based on random basis. All drivers drove the same reference route and vehicle as described in Section 5.1.

### 6.3 Case Study

Along the reference route, there are various curves with different radius. The highest curvature is measured in an S-curve combination. This curve is analyzed as a case study.

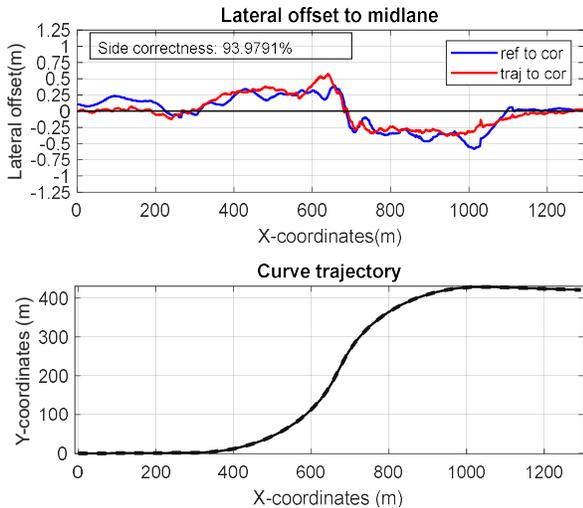

**Fig.8** Case study in an S-curve on country road. Comparing the lateral offset to the mid-lane of the planned ('traj') and the reference ('ref') path curvatures to the corridor ('cor')

We can see the reference and the planned offsets to the mid-lane in Fig.8. In the curve approach phase, the curvature is close to zero, therefore the planned path is close to the mid-lane. In the curve, the planned offset follows the reference with high accuracy. The characteristic of the planned path is the same as that of the reference.

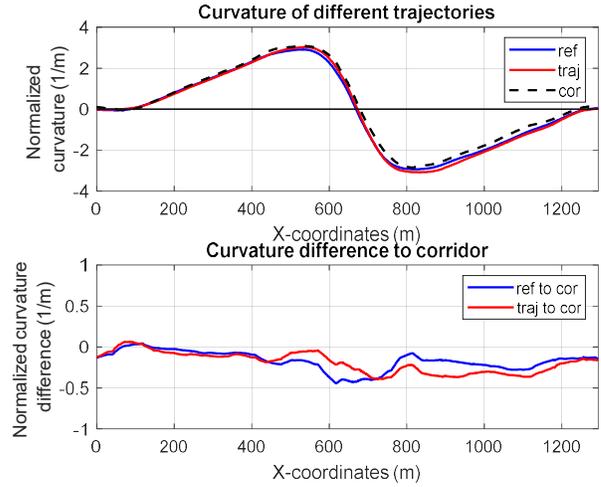

**Fig.9** Case study in an S-curve on country road. Curvature comparison of the planned ('traj') and the reference ('ref') path curvatures to the corridor ('cor')

In Fig.9, curvatures of the planned and the reference paths are compared. In the bottom subplot we can see the curvature difference to the corridor. The planned path shows similar characteristics in curvature as the reference. There is a small positive curvature deviation before the curve which means that the driver starts the cornering before the curve entry. This behavior is observable in the planned path curvature, as well. During cornering there is a negative curvature deviation all along. Both the reference and the planned paths follow this behavior. The negative deviation in a left curve means that the driver uses less steering angle than the mid-lane would need. This is possible due to the higher steering angle before the curve entry. At the tip point, the curvature starts to decrease. This is the start of the right curve. The reference and the planned path curvatures are still lower than the corridor curvature, which means the driver pulls the vehicle more to the right and follows higher steering angle than the corridor would require.

Based on this case study, we can state that in the highest curvature S-curve, the model was able to plan a path which is close to the reference data in terms of both lateral offset and curvature. This means the human curve path selection could be reproduced.

### 6.4 Performance Analysis

The statistical performance analysis includes two evaluation aspects:
- safety and
- deviation from the human paths.

To judge the safety of the planned path we calculate number of sample points where the planned path would result in the vehicle leaving the corridor. The ratio between violating points and the total number of sampling points are shown. If there are no violating points, the minimal distance of the vehicle body (edge points) to the corridor border is given, calculated separately for curve segments.

Then, the worst case of the minimum border distances for the entire route is calculated. The lower this distance is, the less safe the path is.

To determine the performance of the planner we calculate the following indicators:
- Euler distance of the planned path points to the human path points (maximum and average).
- Side correctness: number of points where the planned path lies on the same side of the mid-lane as the reference, compared to the total number of sample points.

The performance indicators are only calculated for the curves. The test route is separated to curve segments, and these curve segments are used only for evaluation. Altogether 10 curves per driver have been studied.

Table 1. Safety results of different drivers. The nominal lane width is 3.70 *m*

| Driver ID | Border violation | Minimum border distance |
|---|---|---|
| Driver 1 | 0 % | 0.458 m |
| Driver 2 | 0 % | 0.422 m |
| Driver 3 | 0 % | 0.627 m |
| Driver 4 | 0 % | 0.313 m |
| Driver 5 | 0 % | 0.156 m |
| Driver 6 | 0 % | 0.578 m |
| Driver 7 | 0 % | 0.143 m |
| Driver 8 | 0 % | 0.448 m |
| Driver 9 | 0 % | 0.473 m |
| Driver 10 | 0 % | 0.38 m |
| Driver 11 | 0 % | 0.439 m |
| Driver 12 | 0 % | 0.025 m |
| Driver 13 | 0.19 % | 0 m |
| Driver 14 | 0 % | 0.439 m |
| Driver 15 | 0 % | 0.438 m |

The safety results are displayed in Table 1. For all drivers the model was able to reproduce safe paths, except for Driver 13. Even though no border violation occurs, the planned path for Driver 12 also resulted in a very low minimum border distance, which is also unsafe. For the other drivers, the minimum border distance has always a minimum value which provides a margin for the controller inaccuracies. Driver 12 and 13 must be further analyzed to reveal why the model is not able to plan safe paths.

Table 2. Statistical results of different drivers – performance evaluation. The nominal lane width is 3.70 *m*

| Driver ID | Average Distance | Side correctness |
|---|---|---|
| Driver 1 | 0.0204 m | 66.56 % |
| Driver 2 | 0.0702 m | 59.96 % |
| Driver 3 | 0.0529 m | 52.57 % |
| Driver 4 | 0.0458 m | 52.79 % |
| Driver 5 | 0.0391 m | 55.45 % |
| Driver 6 | 0.0185 m | 64.79 % |
| Driver 7 | 0.0519 m | 56.42 % |
| Driver 8 | 0.0842 m | 52.44 % |
| Driver 9 | 0.0379 m | 59.17 % |
| Driver 10 | 0.0482 m | 45.1 % |
| Driver 11 | 0.023 m | 68.32 % |
| Driver 12 | 0.071 m | 60.78 % |
| Driver 13 | 0.0577 m | 53.28 % |
| Driver 14 | 0.0343 m | 55.4 % |
| Driver 15 | 0.029 m | 53 % |

The performance analysis results can be seen in Table 2. The best seven drivers, who has an average error distance less than 4 *cm*, and more than 55% side correctness are denoted by green. In their cases, the driver model seems to be working properly, and their paths could be reproduced by the model.

Drivers, whose error distance is above 5 *cm*, or their side correctness is below 50% are denoted by red. The paths of these five drivers could not be successfully reproduced by the model. The offset – curvature correlation plot of one of the deviating drivers can be seen in Fig.10. There are two main differences compared to the behavior of our proposed offset model:
- there is an offset for zero curvature, which means they may have a side preference in straight road sections,
- the behavior is different on the left curvature plane that on the right curvature plane, which indicates that drivers may behave differently in right and left curves.

These effects must be analyzed further, and the model may be modified in the future to improve its accuracy.

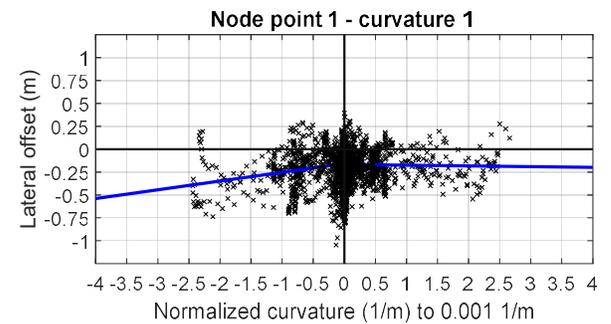

**Fig.10** Correlation plot of Driver 8. $\Delta y_n$ is plotted against $\kappa_{on}$ in each replanning cycle, over the entire test route. Blue solid lines are the linear regression over the left and right curve planes

## 7 Conclusions

In this paper we have aimed to propose a driver model combined with a curve fitting algorithm to produce human-like paths. We have divided the problem into four main parts, the node point model, the offset model, the retrigger model, and the curve fitting model. The perceived environmental information is condensed into node points, and the offset compared to the mid-lane is calculated in these node points.

We propose to use three node points within a look ahead distance of 150 *m,* which provides flexibility to plan various shapes of trajectories, yet avoids high computational complexity. We have analyzed the data of 15 drivers. It has been observed that the road curvature influences the drivers' lane offset selection highly. The relation between road curvature and the selected offset is assumed to be linear. Therefore, we propose a Linear Driver Model, which calculates the node point offsets as the linear combination of the average curvature values between the node points. Three Euler-curves are then fitted onto the node points. This procedure is repeated in even the retrigger cycles. Between cycles, the previously planned path is followed.

We have implemented the algorithm in MATLAB. The parameters of the node point model, the offset model and the retrigger model are given for simulation. The node point distances are given based on our previous work. The concept provides the best fit geometrically on human drivers' path. The resulting node point distances have increasing gaps between each other. The parameter matrix of the offset model is calculated by linear regression onto the measurement data for each driver separately. The retrigger cycle has been selected to provide flexible adaptation on upcoming preview information and providing long replanning cycles to exploit the planned path geometry as much as possible.

All in all, the model performed good in most of the cases. Through a case study, we have shown how a path can be regenerated in curves. Statistically we have proven that the resulting path is close to the human path.

However, there are certain gaps in the results:
- for some drivers the resulting average distance error was too high, or the side correctness of the path was inaccurate,
- for few drivers the resulting path was unsafe.

In the future, we aim to provide further improvement of the model in terms of
- provide a sophisticated node point model, which calculates the node points dynamically based on the environmental information. The effect of improving the node point model on the complete model performance must be quantified.
- Provide an event-based retrigger model.
- Accomplish a trade-off analysis regarding the number of node points.

We consider the results to be satisfying by which we contribute to human-like curve path modelling with a simple, robust solution. In our view, this approach can be well utilized in real time applications in the future.

## Compliance with Ethical Standards

**Conflict of interest** On behalf of all the authors, the corresponding author states that there is no conflict of interest.